\pdfoutput=1

\documentclass[11pt,a4paper]{article}

\usepackage{acl}
\usepackage{times}
\usepackage{latexsym}

\usepackage[T1]{fontenc}

\usepackage[utf8]{inputenc}

\usepackage{microtype}
\usepackage{dsfont}
\usepackage{url}
\usepackage{booktabs} 
\usepackage{graphicx}
\usepackage{tabularx}
\usepackage{amsmath}
\usepackage{bbm}
\usepackage{colortbl}
\usepackage{multirow, makecell, caption}
\usepackage{arydshln}
\usepackage{CJKutf8}

\usepackage{amssymb}
\usepackage{pifont}

\usepackage[normalem]{ulem}
\usepackage{amssymb}
\usepackage{amsfonts}
\usepackage{multicol}
\usepackage{tipa}

\usepackage{csquotes}
\usepackage{xspace}
\usepackage{paralist}
\usepackage{mdwlist}
\usepackage{subfigure}

\usepackage{enumitem}

\newcommand{\ekaremoji}[0]{\includegraphics[height=0.03\textwidth]{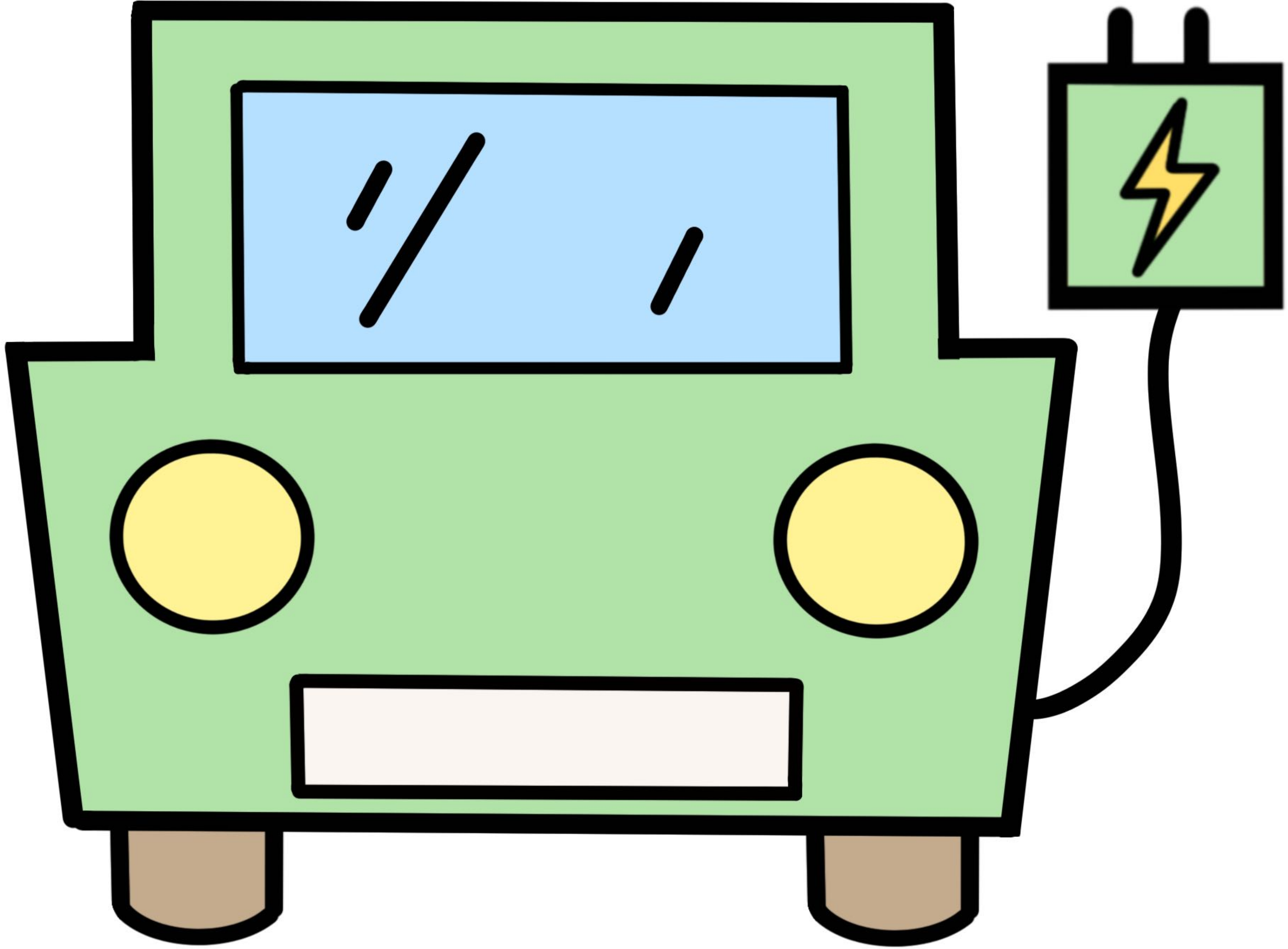}}
\newcommand{\xmark}{\ding{55}}
\newcommand{\cmark}{\ding{51}}
\newcommand{\method}{\textbf{\texttt{E-KAR}}\xspace}

\title{\method\ekaremoji: A Benchmark for Rationalizing Natural Language Analogical Reasoning}

\author{Jiangjie Chen\textsuperscript{\rm $\spadesuit\diamondsuit$}\thanks{~~Work is done during internship at ByteDance AI Lab.}, 
 Rui Xu\textsuperscript{\rm $\spadesuit$}, 
 Ziquan Fu\textsuperscript{\rm $\heartsuit$}, 
 Wei Shi\textsuperscript{\rm $\clubsuit$}, 
 Zhongqiao Li\textsuperscript{\rm $\spadesuit$},\\
 \bf Xinbo Zhang\textsuperscript{\rm $\diamondsuit$},
 Changzhi Sun\textsuperscript{\rm $\diamondsuit$}\thanks{~~Corresponding authors.},
 Lei Li\textsuperscript{\rm $\P$},
 Yanghua Xiao\textsuperscript{\rm $\spadesuit\S$}\footnotemark[2], 
 Hao Zhou\textsuperscript{\rm $\diamondsuit$}\\
\textsuperscript{\rm $\spadesuit$}Shanghai Key Laboratory of Data Science, School of Computer Science, Fudan University\\
\textsuperscript{\rm $\diamondsuit$}ByteDance AI Lab
\textsuperscript{\rm $\heartsuit$}Brain Technologies, Inc.\\
\textsuperscript{\rm $\clubsuit$}South China University of Technology
\textsuperscript{\rm $\P$}University of California Santa Barbara \\
\textsuperscript{\rm $\S$}Fudan-Aishu Cognitive Intelligence Joint Research Center\\
\texttt{\{jjchen19,shawyh\}@fudan.edu.cn},
\texttt{sunchangzhi@bytedance.com}
}

\begin{document}
\begin{CJK}{UTF8}{gbsn}

\maketitle

\begin{abstract}

The ability to recognize analogies is fundamental to human cognition.
Existing benchmarks to test word analogy do not reveal the underneath process of analogical reasoning of neural models.
Holding the belief that models capable of reasoning should be right for the right reasons, we propose a first-of-its-kind Explainable Knowledge-intensive Analogical Reasoning benchmark (\method).
Our benchmark consists of 1,655 (in Chinese) and 1,251 (in English) problems sourced from the Civil Service Exams, which require intensive background knowledge to solve.
More importantly, we design a free-text explanation scheme to explain whether an analogy should be drawn, and manually annotate them for each and every question and candidate answer.
Empirical results suggest that this benchmark is very challenging for some state-of-the-art models for both explanation generation and analogical question answering tasks, which invites further research in this area. 
Project page of \method can be found at \url{https://ekar-leaderboard.github.io}.

\end{abstract}

\section{Introduction}
\label{sec:intro}

\begin{figure}[t]
    \centering
    \includegraphics[width=\linewidth]{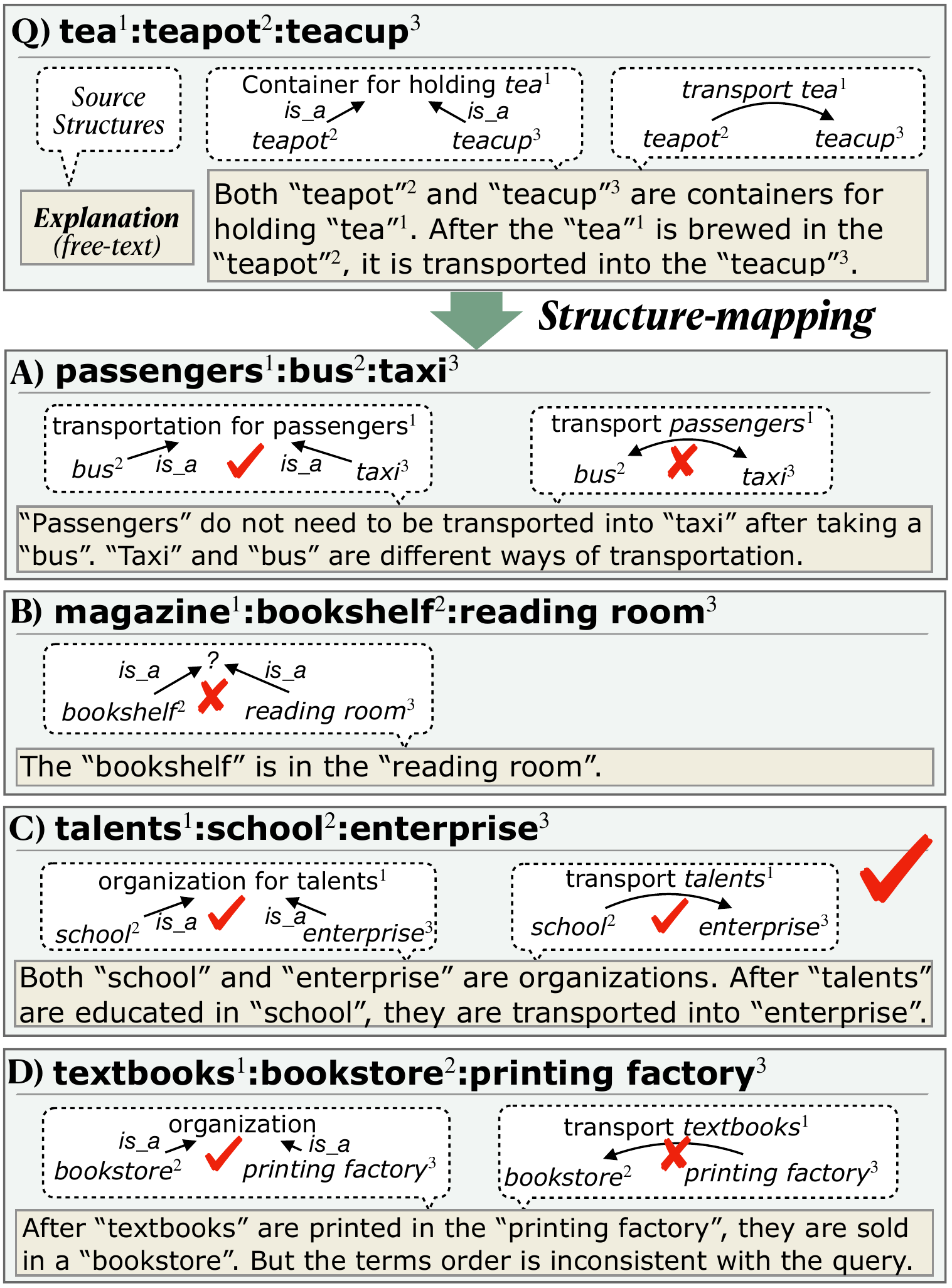}
    \caption{An example in \method. The explanations in \method explain the \textit{structure-mapping} process for analogical reasoning, where source structures are drawn from the query and mapped onto each candidate answer for decision-making.}
    \label{fig:front}
\end{figure}

Analogy holds a vital place in human cognition, driving the discovery of new insights and the justification of everyday reasoning \cite{johnson2006we,gentner2012analogical,bartha2013analogy,10.1145/3448250}.
Due to their unique value in many fields such as creativity \cite{goel1997design} and education \cite{thagard1992analogy}, analogy and analogical reasoning have become a focus in AI research.
The grand question is, are artificial neural networks also capable of recognizing analogies?

Relatively little attention has been paid in NLP to answer this question.
The problem of recognizing analogies is mainly benchmarked in the form of (A:B::C:D) \cite{turney2003combining,mikolov-etal-2013-linguistic,gladkova-etal-2016-analogy,li-etal-2018-analogical} and targeted for testing the ability of pre-trained word embeddings.
Given a tuple of terms as \textit{query} (e.g., \texttt{tea}:\texttt{teapot}:\texttt{teacup}) and a list of \textit{candidate answers} as in Figure \ref{fig:front}, a model needs to find the most analogous candidate to the query, which is C in the example since it matches the relations inherent in the query better than others.

Most methods \cite{mikolov2013distributed,levy-goldberg-2014-linguistic,pennington2014glove} hold a connectionist assumption \cite{feldman1982connectionist} of \emph{linear analogy} \cite{ethayarajh-etal-2019-towards}, that the relation between two words can be estimated by vector arithmetic of word embeddings.
For example, $\vec{\texttt{king}} - \vec{\texttt{man}} + \vec{\texttt{woman}} = \vec{\texttt{queen}}$.
However, current benchmarks focus on the recognition of binary analogies such as syntactic, morphological and direct semantic (e.g., \textit{is\_a} and \textit{synonym\_of}) relations.
And the analogical reasoning procedure behind them is far beyond the scope of this line of research.

In addition, how to explain and rationalize analogical reasoning remains to be the major challenge.
Psychological literature \cite{gick1983schema,gentner1983structure,minnameier2010abduction} suggests that analogical reasoning follows the \emph{structure-mapping} process.
That is, a target (the domain where a problem must be solved, i.e., candidates) and a source (the domain where the analogy is drawn, i.e., the query) are matched, and the relevant features of the source have to be mapped onto the target.
In Figure \ref{fig:front}, source structures are drawn (or \textit{abduced}) from the query and mapped onto candidates, and candidates A, B, D all fail at certain structures.
We argue that such a process can be verbalized into natural language to explain analogical reasoning.


Moving from simply recognizing analogies to exploring human-like reasoning for neural models, we emphasize the importance of a new kind of analogical reasoning benchmark.
To fill in this blank, we propose a first-of-its-kind benchmark for \textbf{E}xplainable \textbf{K}nowledge-intensive \textbf{A}nalogical \textbf{R}easoning (\method).
We collect 1,655 analogical reasoning problems sourced from the publicly available Civil Service Examinations (CSE) of China.
These CSE problems are challenging multiple-choice problems designed by human experts, thus solving them requires the intensive involvement of linguistic, commonsense, encyclopedic, and cultural (e.g., idiom and historical) knowledge.

To justify the reasoning process, we follow the aforementioned guidelines from psychological theories and manually annotate free-text explanations for each query and candidate answers in \method.
Since the annotation requires intensive involvement of knowledge and reasoning, we carefully design a \textit{double-check} procedure for quality control.
We also translate this dataset into an English version, resulting in 1,251 problems after discarding language and cultural specific cases.

In summary, our contributions include:
\begin{itemize}[noitemsep]
    \item We advance the traditional setting of word analogy recognition by introducing a knowledge-intensive analogical reasoning benchmark (\method) in Chinese and English, which is first-of-its-kind and challenging.
    \item To justify the analogical reasoning process, we design free-text explanations according to theories on human cognition, and manually annotate them.
    \item In \method, we define two tasks (analogical QA and explanation generation) in two modes (EASY and HARD) and report the performance of some state-of-the-art language models.
    We discuss the potentials of this benchmark and hope it facilitates future research on analogical reasoning.
\end{itemize}

\section{Related Work}
\label{sec:related}
\paragraph{Word Analogy Recognition in NLP}

Benchmarks for word analogy recognition \cite{turney2003combining,mikolov-etal-2013-linguistic,gladkova-etal-2016-analogy,li-etal-2018-analogical}  examine mostly linear relations between words \cite{ethayarajh-etal-2019-towards}.
Such analogies can often be effectively solved by vector arithmetic for neural word embeddings, such as Word2Vec \cite{mikolov2013distributed} and GloVe \cite{pennington2014glove}.
Recent studies \cite{NEURIPS2020_1457c0d6,ushio-etal-2021-bert} also test such ability of pre-trained language models (PLMs) \cite{radford2019language,devlin-etal-2019-bert,NEURIPS2020_1457c0d6} on these benchmarks.
An exceptional benchmark is \citet{li-etal-2020-ca}, where they build a knowledge-enhanced analogy benchmark that leverages word sense definitions in a commonsense knowledge base \cite{ma-shih-2018-extended}.
However, these benchmarks are mainly set up for evaluating learned representations, and few of them ever investigated the analogical reasoning skills for neural models.
Thus, the goal of this work largely differs from this line of research, as we aim to build a knowledge-intensive benchmark to teach neural models analogical reasoning for correct thinking.

\paragraph{Reasoning Benchmarks from Examinations}

There are abundant benchmarks derived from human examinations to facilitate the study of machine reasoning~\cite{clark2016combining,schoenick2017moving}.
For example, RACE~\cite{lai2017race} is collected from the English exams for middle and high school students, focusing on skills of passage summarization and attitude analysis.
ARC~\cite{clark2018think} contains natural, grade-school science questions authored for human tests.
MCQA~\cite{guo2017effective}, GeoSQA~\cite{huang2019geosqa} and GCRC~\cite{tan-etal-2021-gcrc} are sourced from national college entrance exams of China, measuring a comprehensive set of reasoning abilities.
LogiQA~\cite{Liu2020LogiQAAC} consists of logical reading comprehension problems from Civil Service Exams of China, which is also our source of analogical problems.
ReClor~\cite{yu2020reclor} and LR-LSAT~\cite{wang2021lsat}, collected from Law School Admission Test, aim for testing logical reasoning abilities.
In our work, we focus on analogical reasoning skills for machines and additionally equip \method with annotated explanations to rationalize reasoning.

\paragraph{Explainable NLP Datasets}

One of the most prominent objectives in machine reasoning is giving reasons for a prediction.
In current datasets for explainable NLP, such reasons can be categorized into three classes \cite{wiegreffe-marasovic-2021-review}: 
\begin{inparaenum}[\it 1)]
    \item \textit{highlights explanations} \cite{NEURIPS2018_4c7a167b,yang-etal-2018-hotpotqa,thorne-etal-2018-fever,kwiatkowski-etal-2019-natural}, which are subsets of the input elements to explain a prediction, e.g., words or sentences;
    \item \textit{free-text explanations}  \cite{NEURIPS2018_4c7a167b,zellers2019vcr,aggarwal-etal-2021-explanations} that are textual explanations for justification;
    \item \textit{structured explanations} \cite{mihaylov-etal-2018-suit,khot2020qasc,10.24963/ijcai.2020/537,jhamtani-clark-2020-learning,geva2021did}, which are not fully free-text and generally follow certain structures such as a chain of facts.
\end{inparaenum}
The explanations can be utilized to augment \cite{rajani-etal-2019-explain}, supervise \cite{camburu-etal-2020-make} and evaluate \cite{deyoung-etal-2020-eraser} model predictions.
In this work, we phrase analogical reasoning itself as an instance of machine reasoning tasks with free-text rationales, advancing the research on analogical reasoning from the perspectives of data collection.

\section{Explainable Analogical Reasoning}
\label{sec:analogy}

In this work, we consider a classic setting of analogical reasoning within NLP: recognizing word/term analogies.\footnote{Here, ``term'' corresponds to ``word'' in previous analogy benchmarks, but allows for multiple words.}
This task can be formulated as multiple-choice question-answering.
Given a query tuple $Q$ with $k$ (two or three) terms, and $m$ candidate answer tuples $A = \{A_i\}_{i=1}^m$, the goal is to find the most analogous one in the candidates to the query.

We advocate that reasoning is about giving reasons explaining a prediction.
In order to teach machines to analogize as humans do, we draw inspiration from theories in cognitive psychology to design the forms of explanations.

\subsection{Analogical Reasoning: A Psychological Perspective}
\label{sec:theory}
Before designing suitable forms of explanations, we introduce some important theories from cognitive psychology for a better understanding of analogical reasoning.
In the psychological literature, analogical reasoning is described as a \emph{schema-induction} \cite{gick1983schema} or \emph{structure-mapping} \cite{gentner1983structure} process.
\citet{peirce1896lessons} claimed that analogy is a combination of abductive and inductive reasoning.
\citet{minnameier2010abduction} further developed the inferential process of analogy into three steps, which we take as the guidelines for designing explanations:
\begin{enumerate}[noitemsep]
    \item A possibly suitable structure in the source domain is abduced from the target domain, which might also work for the target;
    \item The specific concepts of the source structure have to be replaced by suitable target concepts (by an inductive inference);
    \item The validity of the transformation is judged w.r.t. solving the target problem.
\end{enumerate}
Take Figure \ref{fig:front} for example:
Source structures can be abduced that both term 2 (\texttt{teapot}) and term 3 (\texttt{teacup}) belong to a concept, and term 1 (\texttt{tea}) can be transported from term 2 to term 3.
The mapping naturally reveals the validity, for example, candidate A is wrong because \texttt{passengers} do not follow a unidirectional transportation (i.e., from \texttt{bus} to \texttt{taxi}) but a bidirectional one.

\subsection{Explanations for Analogical Reasoning}
\label{sec:explanations}

Following the above guidelines, the explanations for the analogical reasoning task should also include three parts: 
\begin{enumerate}[noitemsep]
    \item \textit{Abduction}: description of suitable structures for the query;
    \item \textit{Mapping}: how the structure is mapped onto candidates, analogous to template-filling;
    \item \textit{Validation}: justification for the correctness of the counterfactual mapping.
\end{enumerate}
To this end, we define \emph{free-text explanation} for analogical reasoning, which is one of the most expressive and commonly-used explanations \cite{wiegreffe-marasovic-2021-review}.
We ensure the free-text explanations are self-contained, knowledge-rich, and sufficient to solve the problem as a substitute for the original input.

Specifically, for each query ($Q$) and candidate ($A_i$), we define free-text explanations $\mathcal{E}_Q$ and $\mathcal{E}_{A_i}$.
Following the guidelines in $\mathsection$\ref{sec:theory}, $\mathcal{E}_Q$ should describe the best suited inherent structure of a query abduced from the problem.
$\mathcal{E}_{A_i}$ should decide the correctness in mapping the counterfactual $A_i$ into structure expressed in $\mathcal{E}_Q$, while providing facts as support evidence.

\section{The \method Benchmark}
\label{sec:benchmark}

\subsection{Dataset Collection} 
\label{sec:data_collection}

We build our dataset upon the publicly available problems of Civil Service Exams of China (CSE), which is a comprehensive test for candidates' critical thinking and problem-solving abilities.
CSE consists of problems that test various types of reasoning skills, such as graphical reasoning, logical reasoning and comprehension \cite{ijcai2020-501}, analogical reasoning, etc.

We collect in total 1,655 Chinese analogical reasoning problems from CSE over the years, each of them consisting of a query term tuple and \emph{four} candidate answer tuples of terms (as shown in Figure \ref{fig:front}).
One of the prominent features in CSE problems is the intensive involvement of commonsense, encyclopedic, and idiom knowledge.
For example, one needs to be aware of the fact that ``the \texttt{tide} is caused by both \texttt{Lunar gravity} and \texttt{Solar gravity}''.
More importantly, one needs to know a \textit{negated fact} \cite{barker2012being,hossain-etal-2020-analysis,hosseini-etal-2021-understanding} in order to reject a candidate, such as the fact that ``\texttt{husband} is \textit{not} a \texttt{job}'' or ``a \texttt{car} is \textit{not} made of \texttt{tire}s''.
We keep mainly those requiring knowledge and reasoning skills.
The rest is manually removed, such as the ones testing mathematics, morphology, and phonics, as well as the problems with the number of terms larger than three.

\subsection{Manual Annotation of Explanations}
\label{sec:annotation}

We work with a private company for annotating the explanations defined in $\mathsection$\ref{sec:explanations}.
Before annotation starts, we conduct a training session for all annotators to fully understand the requirements and pick the capable ones based on a selection test.
The selected workers are allocated into two teams, a team of explanation constructors and a team of checkers, where the checkers achieves better scores in the test.
All of them are paid above the local minimum wage.
The annotation consists of two stages:
\textit{1)} the construction stage for writing explanations, 
and \textit{2)} the double-check stage for quality control.

\paragraph{Construction}
During annotation, each problem is assigned to a constructor to build five sentences of explanations: one for query and four for candidate answers.
The explanations are required to be:
\begin{inparaenum}[\it 1)]
    \item fluent and factually correct, 
    \item able to solve the problem on their own, and
    \item knowledge-rich.
\end{inparaenum}
To reduce the labeling difficulty, we allow them to use the search engine for querying the Internet.

\paragraph{First-round Checking}
Afterward, a problem with five annotated explanations is fed to a checker for a first-round checking.
The checker decides whether to accept an explanation sentence according to the criteria in the construction stage.
The rejected ones are sent back to the construction team for revision along with reasons to reject, which serve to re-train the construction team.
The process repeats until a batch reaches 90\% accuracy (i.e., decided to be correct according to the checker).
Then, a second-round checking initiates.

\paragraph{Second-round Checking}
A verified batch is presented to authors for double-checking. 
Authors conduct random inspections for 50\% samples of a batch, and unqualified annotations are sent back with reasons to the check team to fine-tune their checking criteria, which in turn regularize the construction team.
The process also repeats until a batch reaches 95\% accuracy.

In the end, the authors manually calibrate every explanation and acquire 1,655 analogical problems and a total number of 8,275 (5$\times$1,655) free-text explanations, with an average of 31.9 Chinese characters per sentence.

\begin{table}[t]
    \centering
    \small
    \begin{tabular}{lcccc}
    \toprule
    \multirow{2}{*}{\textbf{Dataset}} & \multirow{2}{*}{\textbf{Lang.}} & \textbf{Data Size} & \textbf{\# of Terms} & \textbf{Has} \\
        &  & \textbf{(train / val / test)} & \textbf{in Cand.} & \textbf{Expl.} \\
    \midrule
        SAT & En & 0 / 37 / 337 & 2 & \xmark \\
        Google & En & 0 / 50 / 500 & 2 & \xmark \\
        BATS & En & 0 / 199 / 1,799 & 2 & \xmark \\
        \hdashline
        \multirow{4}{*}{\makecell{\method\\\ekaremoji}} & \multirow{2}{*}{Zh} & \multirow{2}{*}{1,155 /165 / 335 } & 2$_{(64.5\%)}$, & \multirow{2}{*}{\cmark} \\
        & & & 3$_{(35.5\%)}$ & \\
        & \multirow{2}{*}{En} & \multirow{2}{*}{870 / 119 / 262} & 2$_{(60.5\%)}$, & \multirow{2}{*}{\cmark} \\
        & & & 3$_{(39.5\%)}$ & \\
    \bottomrule
    \end{tabular}
    \caption{Comparison between \method and previous analogy benchmarks: language, data sizes in different splits, number of terms in a query or candidate answer, and whether the benchmark has explanations.}
    \label{tab:data_stats}
\end{table}
\subsection{Bilingual \method: English and Chinese}

For a broader impact of this work, we also build an English version of \method via translation.

To translate the Chinese \method into English, we ask three Chinese undergraduate students majoring in English to post-edit the machine-translated results of \method by Google.
Besides translation fluency, we also make sure that
\begin{inparaenum}[\it 1)]
    \item terms in options and explanations have the same word stems;
    \item the parts of speech of terms in a query or candidate answer are encouraged to be the same.
\end{inparaenum}

However, in practice, we notice that some samples in the Chinese dataset can not be accurately translated into English, such as ones involving idioms, poems, and other knowledge of Chinese culture.
Such samples could be hard for non-Chinese people and models to understand without culture-specific knowledge.
Therefore, in the English \method, we manually remove or rewrite these samples, resulting in 1,251 problems and 6,255 (5$\times$1,251) explanations that would require mostly commonsense and factual knowledge and reasoning skills that are universal across cultures and languages.
Nevertheless, those removed samples are valid ones, and the cultural knowledge within them could be of unique value to the Chinese NLP community.
Thus, we keep all samples in the Chinese \method to encourage the research of Chinese NLP.

In the end, we have a bilingual \method for rationalizing analogical reasoning.
Both versions of \method are randomly split into training, development, and test set at the ratio of 7:1:2.
The statistics of \method as well as comparison between previous benchmarks are reported in Table \ref{tab:data_stats}, including SAT \cite{turney2003combining}, Google \cite{mikolov-etal-2013-linguistic} and BATS \cite{gladkova-etal-2016-analogy}.
There are 35.5\%/39.5\% problems with three terms in \method, whereas previous ones only consist of two, making \method even more challenging.

\subsection{Shared Tasks in \method}

Given input $\mathcal{X} = (Q, A)$, the ultimate goal is to make the correct choice $\mathcal{Y}$, while producing rational explanations $\mathcal{E} = \{\mathcal{E}_Q, \mathcal{E}_A = \{\mathcal{E}_{A_i}\}_i\}$.
To this end, we define two shared tasks, \emph{multiple-choice question-answering} (QA) and \emph{explanation generation} (EG), for teaching models how to analogize.

Moreover, to reduce the difficulty of this task as well as follow the structure-mapping process (as in $\mathsection$\ref{sec:analogy}), we propose an easier task form of the shared tasks by adding $\mathcal{E}_Q$ into input $\mathcal{X}$.
Next, we will elaborate on these settings.

\paragraph{Task 1: Analogical QA}

The analogical QA task is formulated as $P_\mathrm{QA}(\mathcal{Y}|\mathcal{X})$.
The QA task requires an understanding of the relationship between the query and each of the candidates to find the correct answer.
For evaluation, we directly use the \emph{accuracy} of multiple-choice QA.

Note that all candidates may be related to the query tuple from certain perspectives.
The challenge lies in finding the \emph{most} related one, i.e., to identify the inherent connections and relations between terms in the query and candidates, considering properties such as linguistic features, order of terms, commonsense knowledge, etc.
For example, the error for candidate D in Figure \ref{fig:front} can be attributed to the incorrect term order, though three terms follow similar relations as in the query. 
Hence, the best choice is C.

\paragraph{Task 2: Explanation Generation}
\label{sec:exp_task}

This task aims to produce a \textit{pipelined rationalization} for analogical reasoning, formulated as $P_\mathrm{EG}(\mathcal{E}|\mathcal{X})$.
The generated explanations $\mathcal{E}$ can be further utilized for the analogical QA, i.e., $P_\mathrm{QA}(\mathcal{Y}|\mathcal{X}, \mathcal{E})$.
Note that the EG task does \textit{not} generate post-hoc explanations for the QA task, therefore there will not be any predicted choice labels in the input $\mathcal{X}$.
Rather, it indicates that the model should make implicit label predictions in explanations \cite{wiegreffe-etal-2021-measuring}.
The generated explanations can be directly evaluated the same as text generation tasks.
Or, indirectly, we can follow a pipelined rationalization paradigm and see how generated explanations can help downstream QA tasks.

\paragraph{Task Mode: EASY vs. HARD}

The abduction of \textit{source structure} (query explanation $\mathcal{E}_Q$) is critical but difficult for making rational analogical reasoning.
Therefore, we propose two task modes:
\begin{itemize}[noitemsep]
    \item \textit{HARD mode}: the original setting, where only $Q$ and $A$ are available in $\mathcal{X}$;
    \item \textit{EASY mode}: in addition to $Q$ and $A$, $\mathcal{E}_Q$ is allowed as part of the given input $\mathcal{X}$.
\end{itemize}

Essentially, EASY mode sets a much clearer playground for evaluating a system's ability to validate counterfactuals (as in $\mathsection$\ref{sec:explanations}):
\textit{What if candidate terms follow the structures in the query instead of query terms? Will they hold logically?}
Therefore, we believe it to be an important supplement for \method benchmark.

\section{Methods}
\label{sec:methods}

In this section, we describe the baseline methods in both QA and EG tasks in EASY and HARD modes.
We mainly evaluate some of the state-of-the-art language models for solving tasks in \method.
Some implementation details are reported in Appendix \ref{appendix:impl}.

\subsection{Baselines for Analogical QA}
\label{sec:baseline_qa}

\paragraph{Pre-trained Methods}
As pre-trained-only baselines, we adopt three static word embeddings that have shown their effectiveness in previous analogy tasks: Word2Vec \cite{mikolov2013distributed}, GloVe \cite{pennington2014glove} and FastText \cite{bojanowski2017enriching}.
We also test contextualized embeddings from PLMs, including BERT \cite{devlin-etal-2019-bert} and RoBERTa \cite{liu2019roberta}.
The averaged token representation is taken as the term representation.
A query or a candidate is estimated by the sum of the representations of each term pair, which is represented as the embedding vector differences \cite{hakami2017compositional,ushio-etal-2021-bert}.
The candidate with the highest cosine similarity to the query is chosen as the answer.

\paragraph{Fine-tuned Methods}
We also set up fine-tuned baselines for QA with PLMs (BERT and RoBERTa).
Since previous benchmarks do not have a training set, we only fine-tune the models on their development set.
The query and candidates are respectively \emph{verbalized} into text using simple prompts, and an example prompt can be found in Appendix \ref{appendix:prompt}.
Each candidate is concatenated with the query into one sentence, which is fed into a PLM for contextualized representation learning.
Averaged hidden states are then fed to an MLP layer and a softmax layer for classification.

\paragraph{Human Evaluation}

We ask three students to solve the QA task in \method, who are undergraduate or graduate students and fluent in English and Chinese.
We randomly sample 100 problems from \method of each language. 
Subjects are asked to first solve them in HARD mode then in EASY mode, in order to reveal the change in performance of the same problem when prompted with the query explanation.
The averaged score is reported as the human baseline.

\subsection{Baselines for Explanation Generation}

We formulate the EG task in a Seq2Seq paradigm, instantiated with state-of-the-art pre-trained language models for Seq2Seq tasks, including BART \cite{lewis-etal-2020-bart,shao2021cpt} and T5 \cite{raffel2020exploring,zhang2021mengzi}.

Although the explanation is individually specific to each query and candidate, the generator has to take into account the whole problem for generating with the \emph{best} source structure (as in $\mathsection$\ref{sec:theory}) and thus finding the most analogous candidate.
Similar to fine-tuned methods in QA task, the EG model takes as input the concatenation of the query $Q$ and all candidate answers $A$ (and the query explanation $\mathcal{E}_Q$ if in EASY mode).
Note that in HARD mode, we switch the prefix of input from generating for $Q$ or $A_i$ in order to distinguish between generating explanations for the query or candidate answer.
An example prompt is presented in Appendix \ref{appendix:prompt}.

\paragraph{Evaluation for the EG Task}
\label{sec:baseline_eg}

In HARD mode, both the generated explanations for query $\mathcal{E}_Q$ and candidate answers $\mathcal{E}_A$ should be evaluated.
In EASY mode, since $\mathcal{E}_Q$ is fed into the model as input, only $\mathcal{E}_A$ are required for evaluation.
The generated text can be evaluated with text generation metrics such as ROUGE \cite{lin2004rouge}, BERTScore\footnote{We use the code of BERTScore at \url{https://github.com/Tiiiger/bert\_score}, where English BERTScore is based on a RoBERTa (large) and Chinese one is based on a BERT (base).} \cite{Zhang2020BERTScore}, BLEURT \cite{sellam-etal-2020-bleurt} and MoverScore \cite{zhao-etal-2019-moverscore}.
However, we would like to highlight that great challenges remain for automatically evaluating semantic-rich text generation \cite{celikyilmaz2020evaluation}.

We also follow the pipelined rationalization paradigm and calculate the gain on QA accuracy as a supplement evaluation metric, i.e., the accuracy drop of $P_\mathrm{QA}(\mathcal{Y}|\mathcal{X}, \mathcal{E})$ over $P_\mathrm{QA}(\mathcal{Y}|\mathcal{X}, \mathcal{E}_{\texttt{gold}})$.
This metric is denoted as \textbf{Acc ($\Delta$)}, where Acc is the QA accuracy when including generated explanations $\mathcal{E}$ as input during inference, and $\Delta$ reflects the accuracy drop.
Here we fix a trained QA model $P_\mathrm{QA}(\cdot)$ based on a large-version RoBERTa.
This model is designed to be \textit{different} from the ones in the QA task, as it is fine-tuned by concatenating gold explanations to the corresponding query or candidates as input during training (prompt detail can be found in Appendix \ref{appendix:prompt}).
As an evaluation metric, we alter the input explanations to the model from \textit{gold $\mathcal{E}$} to \textit{generated $\mathcal{E}$}, and see their performance drops over gold.
Note that the query explanation $\mathcal{E}_Q$ is still the input for all settings in EASY mode.

\section{Results and Analysis}
\label{sec:results}

In the experiments, we wish to answer two questions:
\begin{inparaenum}[\it Q1)]
    \item Can models do knowledge-intensive analogical QA?
    \item Can models generate rational reasons for analogical thinking?
\end{inparaenum}

\paragraph{Categorization of Problems}

We first manually categorize the relational types of problems in \method according to a pre-defined schema.
Unlike free text, we are unable to induce a comprehensive set of relations that covers all candidates due to the complexity of CSE problems.
As a result, we carefully assign at least one relation to each query.
To facilitate analysis, we also try to assign relations to each candidate and query \textit{in the development and test set}, ending up covering 76\% of the candidates and 100\% of the queries.

We refer to several sources of word analogy definitions and textbooks for analogy tests (listed in Appendix \ref{appendix:relation_def}), and categorize the relations into five \emph{meta-relations} (as well as their coverage in the test set) and several accompanying \emph{sub-relations}:
\begin{enumerate}[noitemsep]
    \item \textit{Semantic} (R1, 8.36\% for Zh, 4.12\% for En), the similarity or difference in the meaning of terms, including \textit{synonym\_of} and \textit{antonym\_of};
    \item \textit{Extension} (R2, 41.25\% for Zh, 42.30\% for En), the relation between the extension of terms, including \textit{is\_a}, \textit{contradictory\_to}, etc.;
    \item \textit{Intension} (R3, 37.94\% for Zh, 40.21\% for En), terms relate to each other by inherent properties, including \textit{made\_of}, \textit{has\_function}, etc.;
    \item \textit{Grammar} (R4, 6.36\% for Zh, 6.72\% for En), the grammatical relations between terms, including \textit{subject-predicate}, \textit{head-modifier}, etc.;
    \item \textit{Association} (R5, 6.08\% for Zh, 6.65\% for En), logical association between terms, including \textit{result\_of}, \textit{sufficient\_to}, etc.
\end{enumerate}
Complete sub-relations are presented in Appendix \ref{appendix:relation_def}, as well as their definitions and examples.

\subsection{Can models do knowledge-intensive analogical reasoning?}

\begin{table}[t]
    \centering
    \small
    \begin{tabular}{lccccc}
    \toprule
        \multirow{2}{*}{\textbf{Method}} & \multirow{2}{*}{\textbf{SAT}} & \multirow{2}{*}{\textbf{Google}} & \multirow{2}{*}{\textbf{BATS}} & \multicolumn{2}{c}{\method (H/E)} \\
        \cmidrule{5-6}
        & & & & Zh & En \\
    \midrule
    \rowcolor[gray]{0.95} \multicolumn{6}{c}{\textit{Pre-trained Word Embeddings}}\\
        Word2Vec$^\dagger$ & 41.5 & 93.2 & 63.9 & 28.2/- & 25.6/- \\
        GloVe$^\dagger$ & \textbf{47.7} & 96.0 & 67.6 & 30.9/- & 27.8/- \\
        FastText$^\dagger$  & 47.1 & \textbf{96.6} & \textbf{72.0} & \textbf{31.4}/- & \textbf{28.2}/- \\
    \rowcolor[gray]{0.95} \multicolumn{6}{c}{\textit{Pre-trained Language Models}}\\
        BERT$_b^\dagger$ &  32.9 & 80.8 & 61.5 & 34.5/- & 30.4/- \\
        RoBERTa$_b^\dagger$  & 42.4 & 90.8 & 69.7 & 41.7/- & 37.4/- \\
        RoBERTa$_l^\dagger$  & \textbf{45.4} & \textbf{93.4} & \textbf{72.2} & \textbf{44.6}/- & \textbf{39.0}/- \\
    \rowcolor[gray]{0.95} \multicolumn{6}{c}{\textit{Fine-tuned Language Models}}\\
        BERT$_b$ & 38.9 & 86.6 & 68.0 & 41.8/46.7 & 37.9/42.2 \\
        RoBERTa$_b$  & 47.7 & 93.8 & 75.2 & 46.9/51.1 & 42.2/48.1 \\
        RoBERTa$_l$  & \textbf{51.6} & \textbf{96.9} & \textbf{78.2} & \textbf{50.1}/\textbf{54.8} & \textbf{46.7}/\textbf{50.5} \\
    \midrule
    Human & - & - & - & \multicolumn{2}{c}{77.8/83.3} \\

    \bottomrule
    \end{tabular}
    \caption{Accuracy results on previous analogy tasks and the QA task in \method. \method (H/E) denotes HARD or EASY mode of analogical QA.
    Method$^\dagger$ is not tuned. PLM$_b$ or PLM$_l$ denote \textit{base} or \textit{large} version, respectively. 
    }
    \label{tab:analogy_results}
\end{table}

Table \ref{tab:analogy_results} reports the accuracy results of baseline methods on previous analogy tasks and the QA task in \method.

\paragraph{How do machines solve analogical reasoning problems?}

To answer this question based on Table \ref{tab:analogy_results}, the findings can be summarized as:

\textit{1)} We find contextualized word embeddings from PLMs not very competitive against static word embeddings in previous analogy tasks, which is consistent with the findings in \citet{peters-etal-2018-dissecting}.

\textit{2)} In a more knowledge-intensive \method, the opposite conclusion can be made, with PLMs prevailing over static word embeddings.

\textit{3)} Furthermore, performance from contextualized representations can be improved in all tasks through fine-tuning, especially for \method, where accuracy increases by roughly 5 to 6 points.

\textit{4)} When incorporating gold source structure (i.e., EASY mode), the QA results significantly improve by roughly 5 points in both languages.

\textit{5)} Moreover, despite our efforts to eliminate culture-specific samples in English \method, the accuracy still falls behind its Chinese counterpart, which could be attribute to: 
\textit{a)} fewer training samples, 
\textit{b)} language-specific pre-training and 
\textit{c)} language-specific information noise by translation.

\paragraph{How do humans solve analogical reasoning problems?}

In contrast to machines, humans achieve in \method 77.8\% accuracy in HARD mode and 83.3\% in EASY mode, indicating the challenge of this task as well as showing that current SOTA language models still fall far behind human performance.
We also find the trend of human performance is generally aligned with machines, with accuracy boost (also $\sim$5 points) when prompted with query explanations.

\begin{figure}[t]
    \centering
	\subfigure[Meta-relations distributions and their error ratios.] { \label{fig:error1} 
		\includegraphics[width=0.47\linewidth]{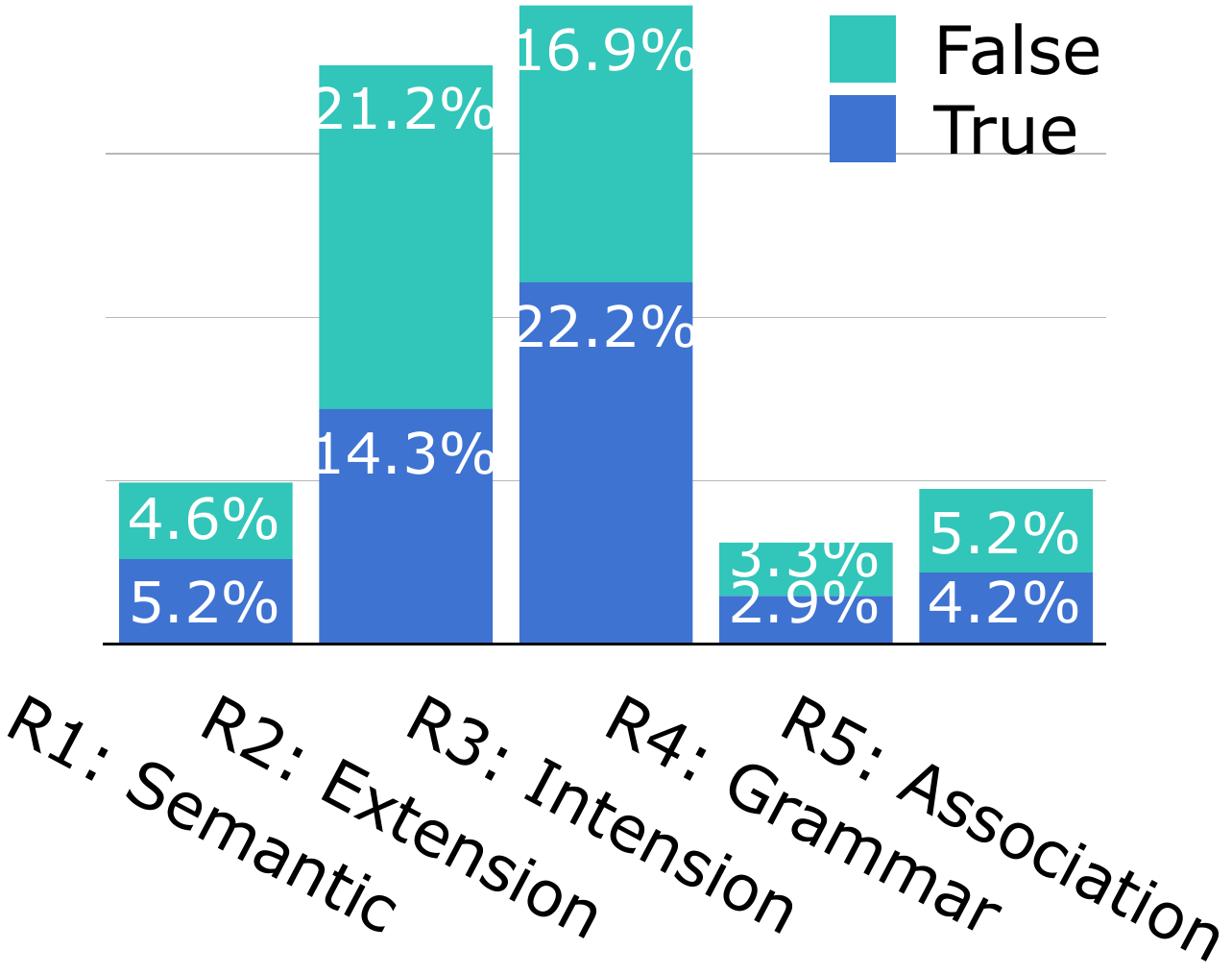}}
	\subfigure[Sub-relations in a sorted order of error rate.] { 
	\label{fig:error2} 
		\includegraphics[width=0.47\linewidth]{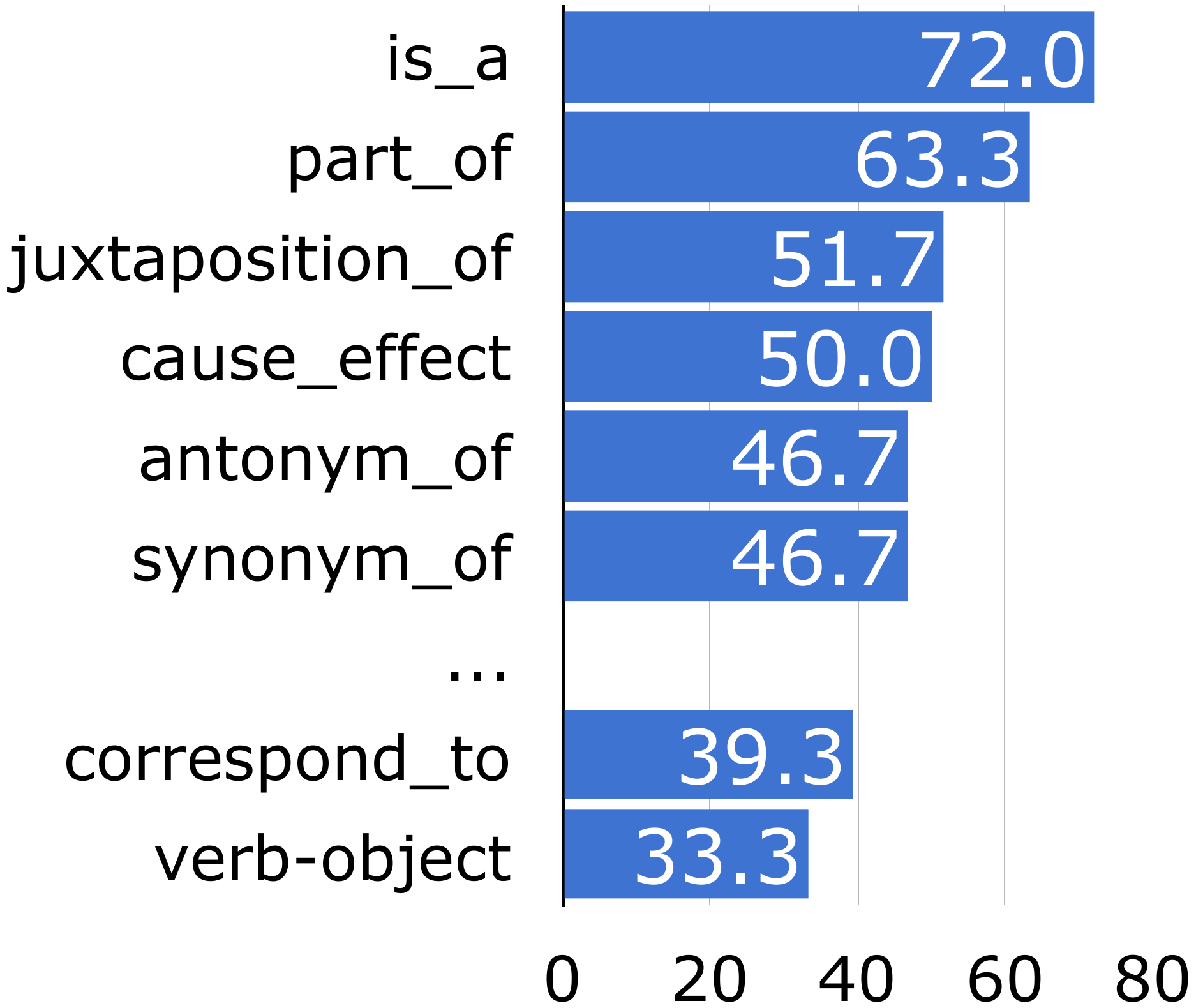}}
    \caption{Error analysis of different query relations. The results are predicted by a fine-tuned RoBERTa (large) in $\mathsection$\ref{sec:baseline_qa} on \method (Zh).}
    \label{fig:errors}
\end{figure}

\begin{table*}[t]
    \centering
    \small
    \begin{tabular}{l||cccc|c||cccc|c}
    \hline
        \multirow{2}{*}{\textbf{EG Method}} &   \multicolumn{5}{c||}{\method (Zh)} & \multicolumn{5}{c}{\method (En)} \\
        \cline{2-11}
        & \textbf{ROUGE} & \textbf{BERT.} & \textbf{BLRT.} & \textbf{Mover.} & \textbf{Acc $\uparrow$ ($\Delta$ $\downarrow$)} & \textbf{ROUGE} & \textbf{BERT.} & \textbf{BLRT.} & \textbf{Mover.} & \textbf{Acc $\uparrow$ ($\Delta$ $\downarrow$)} \\
        \hline 
        \hline 
        None (\xmark) & N/A & N/A & N/A & N/A & 29.1 (68.6) & N/A & N/A & N/A & N/A & 25.6 (72.1) \\
        \hdashline
         BART$_b$ (\xmark) & 39.85 & 72.68 & 63.43 & 64.72 & 33.0
         (64.7) & 17.71 & 91.27 & 54.40 & 59.91 & 29.0 (68.7) \\
         BART$_l$ (\xmark) & 40.39 & 72.67 & 63.60 & 64.57 & \textbf{38.8 (58.9)} & 18.34 & \textbf{91.54} & \textbf{55.48} & 60.13 & \textbf{34.1 (67.6)} \\
         T5$_b$ (\xmark) & \textbf{43.37} & \textbf{83.17} & \textbf{66.34} & \textbf{75.92} & 30.7 (67.0) & 17.44 & 91.17 & 53.71 & 60.40 & 25.6 (72.1) \\
         T5$_l$ (\xmark) & - & - & - & - & - & \textbf{19.77} & 91.44 & 55.00 & \textbf{60.78} & 29.4 (68.3) \\
        \hline
        \hline
         None (\cmark) & N/A & N/A & N/A & N/A & 30.5 (67.2) & N/A & N/A & N/A & N/A & 26.7 (71.0) \\
        \hdashline
         BART$_b$ (\cmark) & 39.08 & 72.84 & 62.10 & 65.07 & 33.4 (64.3) & 25.14 & 91.85 & 56.16 & 62.16 & 29.8 (67.9) \\
         BART$_l$ (\cmark) & 39.18 & 72.93 & 62.45 & 65.13 & \textbf{36.1 (61.6)} & 25.31 & 91.92 & 56.14 & 62.26 & \textbf{32.4 (65.3)} \\
         T5$_b$ (\cmark) & \textbf{40.04} & \textbf{82.52} & \textbf{63.54} & \textbf{74.99} & 34.0 (63.7) & 26.59 & 92.12 & 57.39 & 63.01 & 30.2 (67.5) \\
         T5$_l$ (\cmark) & - & - & - & - & - & \textbf{28.10} & \textbf{92.38} & \textbf{58.76} & \textbf{63.64} & 31.3 (66.4) \\
        \hline
        \hline
        Gold & N/A & N/A & N/A & N/A & 97.7 (0.0) & N/A & N/A & N/A & N/A & 97.7 (0.0) \\
    \hline
    \end{tabular}
    \caption{Results of explanation generation models w.r.t. ROUGE-2, BERTScore, BLEURT, MoverScore and Acc ($\Delta$) on the analogical QA task, where EASY mode (\cmark) incorporates gold $\mathcal{E}_Q$ as part of the model input.
    Note that the QA model here is trained as described in $\mathsection$\ref{sec:baseline_eg}, and we switch input explanations during inference.}
    \label{tab:eg_results}
\end{table*}

\paragraph{Error Analysis for QA}

We further conduct an error analysis based on the results in \method (Zh) predicted by a fine-tuned RoBERTa (large).
The erroneous ones are classified based on the manually annotated meta-relations and sub-relations of \textit{queries}, which is a fine-grained tool for analyzing a model's predictions.

Figure \ref{fig:error1} shows that the model performs poorly on nearly all meta-relations, with R2 (Extension) being the most error-prone one (only 40.3\% accuracy, normalized) and R3 (Intension) being the least one (56.8\% accuracy).
One of the most prominent reasons is that R2 and R3 rely heavily on commonsense and encyclopedic knowledge and reasoning skills such as commonsense and world knowledge, at which current models easily fail.

Figure \ref{fig:error2} shows the error rate of sub-relations with more than 10 samples.
Consistent with Figure \ref{fig:error1}, the three most error-prone sub-relations (\textit{is\_a}, \textit{part\_of} and \textit{juxtaposition\_of}) all belong to R2 (Extension).
Besides, the model seems to do well in linguistic knowledge, with \textit{verb-object} achieving only 33.3\% error rate.
These findings may shed light on future directions for knowledge-intensive reasoning with language models.

\subsection{Can models rationalize analogical thinking?}

We report the automatic evaluation results of generated explanations in Table \ref{tab:eg_results}.
However, such results hardly mean anything due to the incapability to evaluate the semantic-rich text of current automatic metrics.
Therefore, the following analyses mainly focus on Acc ($\Delta$) and human evaluation.

\paragraph{Can (generated) explanations benefit analogical QA?}

To start with, we highlight again that the QA model in Table \ref{tab:eg_results} is different from the one in Table \ref{tab:analogy_results} since the training of the former involves gold explanations.
When exposing gold explanations to the QA model, it achieves 97.7\% accuracy on \method of both languages coincidentally.

However, the QA model performs poorly when removing the explanations during inference (i.e., \textit{None}).
This is because the pipelined rationalization in training makes the QA model rely heavily on the rationales (explanations) than the problem itself, and the removal of them causes severe performance degradation.
When we switch the explanations to generated ones during inference, the accuracy gap ($\Delta$) between gold results slightly narrows, with the gain in EASY mode being more significant than in HARD mode.
To conclude, current SOTA generative language models still fall short of rationalizing analogical reasoning, which would be a challenging but interesting future direction.

\paragraph{Error Analysis for EG}

\begin{table}[t]
    \centering
    \small
    \begin{tabularx}{\linewidth}{|l|X|}
    \hline
        Q) & 氧气 (oxygen):臭氧 (ozone) \\
        \hline
        A) & 盐 (salt):氯化钠 (sodium chloride) \\
        B) & 硫酸 (sulfuric acid):硫 (sulfur) \\
        C) & 石墨 (graphite):金刚石 (diamond) \\
        D) & 石灰水 (lime water):氢氧化钙 (calcium hydroxide) \\
    \hline
    \hline
        \textbf{$\mathcal{E}_Q$}& \uline{氧气}和\uline{臭氧}都只由氧元素组成。Both \uline{oxygen} and \uline{ozone} are made of only the oxygen element. \\
        \hdashline
        \textbf{$\mathcal{E}_Q^\dagger$}& \uline{臭氧}是\uline{氧气}的一种。\uline{Ozone} is a kind of \uline{oxygen}.\\ 
    \hline
        \textbf{$\mathcal{E}_\mathrm{A}$}& \uline{氯化钠}是\uline{盐}的主要成分，\uline{盐}和\uline{氯化钠}不是只由一种元素组成。\uline{Sodium chloride} is the main component of \uline{salt}. Neither \uline{salt} nor \uline{sodium chloride} is made of only one element. \\
        \hdashline
        \textbf{$\mathcal{E}_\mathrm{A}^\dagger$}& \uline{氯化钠}是\uline{盐}的一种。\uline{Sodium chloride} is a kind of \uline{salt}.\\
    \hline
    \end{tabularx}
    \caption{Case study of EG in HARD mode, where $\mathcal{E}_\ast$ is gold and $\mathcal{E}^\dagger_\ast$ is generated by a BART (large).}
    \label{tab:case}
\end{table}

We also randomly select 100 sentences generated by a BART (large) for manual inspection by the authors.
Aside from the common errors in generation models such as repetition, we find that task-specific errors for generated explanations can be roughly categorized into three classes: 
\begin{inparaenum}[\it 1)]
    \item \textit{unable to generate negated facts to refute source structure};
    \item \textit{generating factually incorrect statements};
    \item \textit{biasing towards common patterns}, e.g., ``\texttt{term 1} and \texttt{term 2} have similar meanings'' and ``\texttt{term 1} is a \texttt{term 2}''.
\end{inparaenum}
For example, in Table \ref{tab:case}, both generated $\mathcal{E}_Q$ (only in HARD mode) and $\mathcal{E}_\mathrm{A}$ are factually incorrect, and the model fails to generate the negated fact that ``both are not exclusively made of one component.''

We dig further into the first class of errors (w.r.t. negation), which is important to refute a candidate, as mentioned in $\mathsection$\ref{sec:data_collection}.
We find $\sim$90\% gold explanations of wrong candidates contain negated statements.
Yet, the number drops to 14.9\% (Zh) and 22.1\% (En) in the generated ones in HARD mode, and 21.3\% (Zh) and 38.6\% (En) in EASY mode.
An interesting conclusion can be drawn that current generative models do not seem to know how to generate a negated yet truthful fact, such as ``\texttt{feeling} can \textit{not} guide \texttt{psychological reaction}.'' since feeling \textit{is} a reaction. 
And exposing source structure to the model (EASY mode) seems to alleviate this problem.

The fact also questions the astonishing QA performance by adding gold explanations (97.7\%), as the model could be biased towards surface-level negation.
To debias this, we conduct a simple ablation study by directly removing the clauses containing the negation word ``不'' (\textit{not}) from the gold explanations in the test set, and still achieve 92.5\% in QA accuracy.
This finding indicates that the QA model with correct rationales would not be very much biased towards negation in the explanation.

\section{Conclusion and Discussion}
\label{sec:conclusion}

In this work, we propose a first-of-its-kind benchmark \method (in both Chinese and English) for explainable analogical reasoning, which sets a concrete playground and evaluation benchmark to boost the development of human-like analogical reasoning algorithms.
The \method benchmark is featured by its rich coverage in knowledge and well-designed free-text explanations to rationalize the analogical reasoning process.
Preliminary experiments show that this benchmark provides a rather difficult challenge for prevailing language models.

However, there are still many open questions to be addressed.
For example, humans solve the analogy problems in a trial-and-error manner, i.e., adjusting the abduced source structure and trying to find the most suited one for all candidate answers.
However, the explanation annotation process in \method (not the EG task) is mostly post-hoc and reflects only the result of reasoning.
Such explanations cannot offer supervision for intermediate reasoning, though it is an interesting question whether an intelligent model should be deeply supervised at every step \cite{tafjord-etal-2021-proofwriter}.
Furthermore, \method only presents one feasible explanation for each problem, whereas there may be several.

This benchmark also invites reasoning models that can effectively interact with extra knowledge.
It remains to be a great challenge to generate and evaluate factually correct explanation text.
Especially, how to generate negated facts is relatively under-explored in the research community but of much importance.
Finally, whether the analogical QA system can \textit{correctly} exploit explanations and background knowledge is also worth investigating, which may intersect with research on debiasing \cite{NEURIPS2020_1091660f,niu2021counterfactual}.

We hope this work to be a valuable supplement to future research on natural language reasoning, especially for research on analogical reasoning and explainable NLP.

\section*{Acknowledgement}
We thank the anonymous reviewers for their valuable suggestions.
We also thank Ruxin Yu for the logo design.
This work was supported by National Key Research and Development Project (No. 2020AAA0109302), Shanghai Science and Technology Innovation Action Plan (No.19511120400) and Shanghai Municipal Science and Technology Major Project (No.2021SHZDZX0103).

\clearpage
\section*{Ethical Considerations}
\label{sec:ethic}

This paper proposes a new kind of analogical benchmark with explanations to rationalize models' predictions.
The dataset is collected from Civil Service Exams of China, which is publicly available and has been used in other public datasets before, such as LogiQA \cite{Liu2020LogiQAAC}.
The annotated explanations for each problem in our dataset are crowd-sourced by working with ByteDance.
The construction team remains anonymous to the authors, and the annotation quality is guaranteed by the double-check strategy as mentioned in $\mathsection$\ref{sec:annotation}.
We ensure that all annotators' privacy rights are respected in the annotation process.
All annotators have been paid above local minimum wage and consented to use the datasets for research purposes covered in our paper.

\bibliography{custom}
\bibliographystyle{acl_natbib}

\clearpage
\begin{appendix}
\label{sec:appendix}

\section{Implementation Details}
\label{appendix:impl}

The pre-trained word embeddings are provided by \citet{P18-2023}, and the checkpoints for PLMs are hosted in HuggingFace \cite{wolf-etal-2020-transformers}.
Most of the parameters in the baseline models take the default values from HuggingFace's Transformers library, and we keep the best checkpoint on the validation set for testing.
The Chinese version of BERT (whole word masking) and RoBERTa (whole word masking extended) are provided by \citet{cui-etal-2020-revisiting}, BART by \citet{shao2021cpt} and T5 by \citet{zhang2021mengzi}.\footnote{Note that the Chinese T5 (Mengzi) does not have large version, as they claim to be lightweight but ingenious.}
Thus the EG results of T5 in \method (zh) can be attributed to both \citet{raffel2020exploring} and \citet{zhang2021mengzi}.

\subsection{Example Prompts in \method}
\label{appendix:prompt}

We denote terms in a query $Q$ or a candidate $A^\ast\in \{A, B, C, D\}$ as $t_{Q/A^\ast}^{\{1,2\}}$.
The example prompts for the QA and EG tasks in \method are:
\begin{itemize}
    \item \textit{A Prompt for the QA Task}: ``\texttt{(context: $\mathcal{E}_\ast$,) question: $t_Q^1:t_Q^2$, options: $t_A^1:t_A^2$, $t_B^1:t_B^2$, $\cdots$ or $t_D^1:t_D^2$}''.
    \item \textit{A Prompt for the EG Task}: ``\texttt{query = $t_Q^1:t_Q^2$ </s> (query explanation = $\mathcal{E}_Q$) </s> candidate = $t_A^1:t_A^2$ </s> candidate = $t_B^1:t_B^2$ </s> $\cdots$ </s> candidate = $t_D^1:t_D^2$ </s> generate the explanation of $Q$/$A_i$:}''.
    \item \textit{A Prompt for the QA model in Acc $\Delta$}: concatenating explanations to the query and each candidate answer, such as  ``\texttt{$t_Q^1:t_Q^2$ </s> explanation: $\mathcal{E}_Q$}'' and ``\texttt{$t_A^1:t_A^2$ </s> explanation: $\mathcal{E}_A$}''.
\end{itemize}

\section{Detailed Relation Definitions}
\label{appendix:relation_def}

To design the relation taxonomy, we refer to a number of sources that categorize types of analogy tests, including MAT\footnote{http://www.west.net/\~stewart/mat/analogies\_types.htm}, Fibonicci\footnote{https://www.fibonicci.com/verbal-reasoning/analogies-examples/}, Offcn Education (in Chinese)\footnote{https://www.offcn.com} and Huatu Education (in Chinese)\footnote{https://www.huatu.com}, etc.

The complete set of meta-relations and sub-relations are presented in Table \ref{tab:subrelations}.

\begin{table*}[t]
  \centering
  \small
  \begin{tabularx}{\textwidth}{lXlcc}
  \toprule
    \multirow{2}{*}{\textbf{Relation}} & \multirow{2}{*}{\textbf{Definition}} & \multirow{2}{*}{\textbf{Example}} & \multicolumn{2}{c}{\textbf{Coverage}} \\
    \cmidrule{4-5}
    & & & Zh & En \\
  \midrule 
  
  \rowcolor[gray]{0.95} \multicolumn{3}{l}{\textbf{R1: \textit{Semantic}}}  & \textbf{8.36\%} & \textbf{4.12\%} \\
    1) \textit{synonym\_of} & The meanings of two terms are similar. & clarity : transparency & 4.88\% & 2.37\% \\
     \hdashline 
    2) \textit{antonym\_of} & The meaning of two terms are opposite or used to express different concepts.  & harmony : conflict & 3.48\% & 1.75\% \\
    \midrule
    
    \rowcolor[gray]{0.95} \multicolumn{3}{l}{\textbf{R2: \textit{Extension}}}  & \textbf{41.25\%} & \textbf{42.30\%} \\
    1) \textit{identical\_to}  & The meanings of two terms are identical. & highway : road & 1.64\% & 0.92\% \\
     \hdashline 
    2) \textit{is\_a} & One term is the hypernym of the other. & Earth : planet & 11.54\% & 12.38\% \\
     \hdashline 
    3) \textit{part\_of} & One term is a part of the other. & steering wheel : sedan & 6.82\% & 7.78\% \\
     \hdashline 
    4) \textit{juxtaposition\_to} & Two terms belong to the same hypernym or have the same properties or functions. & shoes : socks & 12.86\% & 12.62\% \\
     \hdashline 
    5) \textit{contradictory\_to} & Two term are contradictory to each other. & vowel : consonant & 1.19\% & 1.25\% \\
     \hdashline 
    6) \textit{contrary\_to} & Two propositions cannot both be true, but can both be false. & black : white & 4.36\% & 4.08\% \\
     \hdashline 
    7) \textit{intersection\_to} &The extension of the two terms intersects. & solo : pianolude & 2.45\% & 2.81\% \\
     \hdashline 
    8) \textit{utterly\_different} & The extensions of terms do not overlap. & apple : nuts & 0.39\% & 0.46\% \\
    \midrule
    
  \rowcolor[gray]{0.95} \multicolumn{3}{l}{\textbf{R3: \textit{Intension}}}  & \textbf{37.94\%} & \textbf{40.21\%} \\
    1) \textit{attribute\_of} & One term is the attribute of the other. & object : inertia & 1.15\% & 1.17\% \\
     \hdashline 
    2) \textit{probabilistic\_attribute} & One term is probably the attribute of the other. & shoes : high heels & 0.33\% & 0.34\% \\
     \hdashline 
    3) \textit{has\_function} & One term has the function of the other. & calculator : calculate & 2.94\% & 3.54\% \\
     \hdashline 
    4) \textit{metaphor} & A term is the metaphor of the other, reflecting something abstract indirectly.  & pigeon : peace & 1.15\% & 0.42\% \\
     \hdashline 
    5) \textit{takes\_place\_in} & A term takes place in the other. & soldier : battlefield & 0.96\% & 1.07\% \\
     \hdashline 
    6) \textit{located\_in} & A term is located in the other. & Rhine : Europe & 2.06\% & 2.47\% \\
     \hdashline 
    7) \textit{made\_of} &  One term is the raw material of the other. & door : wood & 3.21\% & 3.90\% \\
     \hdashline 
    8) \textit{tool\_of} & One term is the tool of the other. & knives : murder & 0.91\% & 1.00\% \\
     \hdashline 
    9) \textit{target\_of} & One term is the target of the other. & health : exercise & 0.82\% & 0.72\% \\
     \hdashline
    10) \textit{corresponds\_to} & Terms generally correspond to each other. & post office : mail bank & 24.41\% & 25.58\% \\
    \midrule
    
  \rowcolor[gray]{0.95} \multicolumn{3}{l}{\textbf{R4: \textit{Grammar}}}  & \textbf{6.36\%} & \textbf{6.72\%} \\
    1) \textit{subject-predicate} & The originator of the action and the action itself.  & plane : take off & 1.19\% & 1.25\% \\
     \hdashline 
    2) \textit{verb-object} & The action and the object on which the action acts. & transfer : goods & 3.14\% & 3.36\% \\
     \hdashline 
    3) \textit{head-modifier} & The preceding term modifies the other. & affluence : living & 0.87\% & 0.74\% \\
     \hdashline 
    4) \textit{subject-object} & The originator and receiver of an action. & dairy farmer : milk & 1.16\% & 1.37\% \\
    \midrule
    
  \rowcolor[gray]{0.95} \multicolumn{3}{l}{\textbf{R5: \textit{Association}}} & \textbf{6.08\%} & \textbf{6.65\%} \\
    1) \textit{result\_of} & One term causes the other. & lack of water : plants wither & 2.99\% & 2.97\% \\
     \hdashline 
    2) \textit{follow} & The terms have a chronological or other sequential relationship, but one term does not cause the other. & sign up : take the exam  & 1.91\% & 2.19\% \\
     \hdashline 
    3) \textit{sufficient\_to} & One term is a sufficient condition for the other. & raining : wet ground & 0.0\% & 0.0\% \\
     \hdashline 
    4) \textit{necessary\_to} & One term is a necessary condition for the other. & admission : graduation & 1.18\% & 1.49\% \\
    
  \bottomrule
  \end{tabularx}
  \caption{Complete set of defined sub-relations with definitions, examples and coverage in the test set of \method.}
  \label{tab:subrelations}
\end{table*}

\end{appendix}

\end{CJK}
\end{document}